%% file: main.tex
\documentclass{article}

\RequirePackage{hyperref} 
\RequirePackage{graphicx}
\RequirePackage{amsmath}
\RequirePackage[noblocks]{authblk}
\RequirePackage{amssymb}
\RequirePackage{fullpage}
\RequirePackage{fancyhdr}

\usepackage{makecell}
\usepackage{comment}
\usepackage{multirow}
\usepackage{longtable}
\usepackage{subcaption}
\usepackage{xspace}
\usepackage{xcolor}
\usepackage{wrapfig}
\usepackage{afterpage}

\graphicspath{./Figures}

\let\oldnl\nl

\newcommand{\nonl}{\renewcommand{\nl}{\let\nl\oldnl}}
\newcommand{\cmmnt}[1]{\ignorespaces}  

\newcommand{\HRule}{\rule{\linewidth}{0.5mm}}

\makeatletter
\renewcommand{\@maketitle}{%
  \parindent=0pt
  \centering
  {\Large \bfseries\textsc{\@title}}
  \HRule\par%
  \textit{\@author \hfill}
  \par
}

\makeatletter
\def\thickhline{%
  \noalign{\ifnum0=`}\fi\hrule \@height \thickarrayrulewidth \futurelet
   \reserved@a\@xthickhline}
\def\@xthickhline{\ifx\reserved@a\thickhline
               \vskip\doublerulesep
               \vskip-\thickarrayrulewidth
             \fi
      \ifnum0=`{\fi}}
\makeatother

\newlength{\thickarrayrulewidth}
\makeatother 

\providecommand{\keywords}[1]{\textbf{\textit{Keywords:}} #1}

\begin{document}
\sloppy

\title{A Benchmark Study on Time Series Clustering}
\author[1]{Ali Javed} 
\author[1]{Byung Suk Lee}
\author[2]{Donna M. Rizzo}

\affil[1]{Department of Computer Science, University of Vermont, Burlington, VT, USA}
\affil[2]{Department of Civil \& Environmental Engineering, University of Vermont, Burlington, VT, USA}

\maketitle 

\bigskip

    

\begin{abstract}
This paper presents the first time series clustering benchmark utilizing all time series datasets currently available in the University of California Riverside (UCR) archive --- the state of the art repository of time series data. Specifically, the benchmark examines eight popular clustering methods representing three categories of clustering algorithms (partitional, hierarchical and density-based) and three types of distance measures (Euclidean, dynamic time warping, and shape-based). We frame and follow six restrictions with special attention to making the benchmark as unbiased as possible. A phased evaluation approach was then designed for summarizing dataset-level assessment metrics and discussing the results. The  benchmark study presented can be a useful reference for the research community on its own; and the dataset-level assessment metrics reported may be used for designing evaluation frameworks to answer different research questions.
\end{abstract}



\input{Introduction}
\input{RelatedWork}

\input{Methods}

\input{Results}
\input{Discussion}

\input{Conclusion}
\input{Acknowledgements}

\bibliographystyle{ieeetr}
\bibliography{main.bib}

\input{Appendix}

\end{document}

%% file: Introduction.tex
\section{Introduction}\label{sec:Introduction}

A time series is a sequence of variable values ordered by time. Time series data abound in various disciplines. Examples include stock market ticker price data~\cite{AGHABOZORGI2015}, gene expression sequence data~\cite{AGHABOZORGI2015}, sensor-generated data, CPU usage data, and network monitoring data.  These data are analyzed using a variety of statistical techniques, such as classification, clustering, and anomaly detection. This paper focuses on clustering. Clustering is a well-known unsupervised machine learning method for dividing data points (i.e., observations) into groups (called ``clusters'') such that observations within the same cluster tend to be more similar (according to a pre-specified criteria) than those in different clusters~\cite{Wubook}. 

With the increasing prevalence of time series data, time series clustering has been gaining much attention over the past decade in order to identify previously unknown trends ~\cite{Paparrizos2016,Paparrizos2017,densitypeaks,Nurjahan2016}. The evaluation of clustering algorithms, however, is inherently challenging because these statistical algorithms are, by design, exploratory in nature. For this reason, the algorithm evaluation must rely on empirical study, essentially assessing how well the algorithm ``rediscovers'' already known classifications~\cite{Paparrizos2016,Paparrizos2017,Nurjahan2016} of a given time series data. 

The University of California (UCR) time series archive~\cite{UCRArchive2018} is arguably the most popular and largest labeled time series data archive, with thousands of citations and downloads. At the time of this writing, the archive had a total of 128 datasets comprising a variety of synthetic, real, raw and pre-processed data. The archive was originally born out of frustration, with \emph{classification} research papers reporting error rates on a single time series dataset and implying that the results would generalize to other datasets.
In order to standardize the evaluation of algorithms, each dataset in the UCR archive has been split into training and test data. Additionally, each dataset is accompanied by three baseline straw man classification accuracy scores obtained using the K-nearest neighbor algorithm and different input parameter settings (window size) for dynamic time warping (DTW)~\cite{Sakoe1978}.

Despite extensive use of the archive in creating, validating and evaluating some of the most recently popular time series clustering algorithms~\cite{Paparrizos2016,Paparrizos2017,Nurjahan2016}, at the time of this writing, the archive provides no equivalent \emph{assessment metrics} for assisting with evaluation or validation of the clustering algorithms. The latter is the single largest limitation of the archive when used for assessing clustering algorithms. Different researchers must repeat the process of implementing and benchmarking clustering algorithms over the same data sets. At a minimum, this may cost months or longer of run time~\cite{Paparrizos2017}; and when benchmark tests are repeated, the subjective nature of test details (e.g., pre-processing) may introduce bias that affects the objectivity and re-producibility of the test results.

The work presented in this paper aims to address the limitation associated with testing time series clustering algorithms by providing a clustering benchmark. The intent of this benchmark is similar to the classification benchmark \cmmnt{ accompanying the archive} of Dau et al. (2018)~\cite{UCRArchive2018}, that is to provide comparison with several established methods in order to reduce both the repetition of experiments and time to publication. We would add to this another goal, that is to study the impact of changing design choices that occur within a given clustering method (i.e., a combination of clustering algorithm and distance measure). Additionally, the discussion highlights the value of considering a pool of clustering methods for use in cluster analysis and provides guidance on how to select individual algorithms in such a pool.
To this end, we select eight clustering methods in this benchmark study that span three types of clustering algorithms and three distance measures, and assess each while adhering to the six restrictions laid out below. 
\begin{enumerate}
    \item \label{enum:no_preprocessing} \emph{No pre-processing.} All datasets in the archive were used without any additional pre-processing (e.g., normalization in magnitude, filtering, smoothing). The reason is that, while pre-processing is common and is shown to improve results~\cite{Rakthanmanon2012}, any improvement resulting from the pre-processing should not be attributed to the clustering method itself~\cite{UCRArchive2018,Keogh2003} and, even if it were, the same pre-processing may have different performance impacts on different clustering methods. (About 80\% of the UCR datasets are already z-normalized.)
    \item \emph{Only uniform length time series.} Only datasets in which all time series have equal length are used. The reason is that some of the clustering methods used in this benchmark were designed to work only with time series of equal length. (Only 11 out of 128 datasets in the archive have varying time series length.)
    \item \label{enum:gp_knc}\emph{Known number of clusters.} The clustering methods used in this work require that the number of clusters, $k$, be provided as input. The value of $k$ is known from the class labels annotated in the datasets. There are several techniques for estimating $k$~\cite{Bholowalia2014, Patil2019, SUBBALAKSHMI2015,Bezdek1998}, but evaluating those techniques is not part of this benchmark.
    \item \emph{Minimum two classes.} Only datasets with $k$ = 2 or more classes (other than a class designated as ``noise'') are used, as clustering time series data that all belong to the same class (i.e., $k = 1$) is not meaningful. (Five datasets have less than two classes.) 
    \item\label{enum:established} \emph{Established methods.} All clustering methods used in this work are well-established or have survived the test of time. They are treated with equal merit with no effort to identify one as ``superior'' or ``inferior'' to another.
    \item\label{enum:gp_raw} \emph{Dataset-level assessment metrics.} The assessment metrics are reported for each clustering method on each of the 112 remaining datasets. Using assessment metrics at the dataset level enables evaluation frameworks to be designed with the research questions in mind, eliminating repetitive experimentation.

\end{enumerate}

The remainder of the paper is organized as follows. Section~\ref{sec:relatedwork} discusses related work. Section~\ref{sec:benchmark_methods} describes the benchmark methods. Section~\ref{sec:results} presents the benchmark test results. Section~\ref{sec:conclusion} concludes the paper.

%% file: RelatedWork.tex
\section{Related work}\label{sec:relatedwork}

Benchmarking, in general, has been recognized as an important step in advancing the knowledge of both supervised and unsupervised learning~\cite{Keogh2003,Boulesteix2018,Hui2010,Franti2018}. See Keogh and Kasetty~\cite{Keogh2003} for a nice summary on the need to benchmark time series algorithms. They highlight many studies that use straw man algorithms to compare time series classification algorithms, and note that many of these algorithms provide little value because the levels of improvement are completely dwarfed by the variance observed when tested on real datasets or when minor unstated implementation details change. After a thorough survey of more than 350 time series data mining papers, they concluded that a median of only 1.0 (or an average of 0.91) rival methods were compared against a ``novel'' method (e.g., clustering algorithm, distance measure, pre-processing); and on average, each method was tested on only 1.85 datasets.  While their summary is based on time series \emph{classification}, the same concerns apply to time series \emph{clustering}. 

Works that compare time series clustering methods suggest that these comparisons have either been done qualitatively, using a theoretical approach~\cite{WARREN2005, Ali2019,Roddick2002}, or quantitatively using an empirical approach\cmmnt{ (e.g., via an accuracy metric like the Rand index scores)}~\cite{Paparrizos2016,Paparrizos2017,Nurjahan2016}. 
Only the empirical approaches provide evidence of performance measured on external datasets. The UCR archive has been used for that purpose in most of the recent time series clustering comparisons~\cite{Paparrizos2016,Paparrizos2017,Nurjahan2016}. 
However, none of them reports assessment metrics at the dataset level accounting for all datasets in the archive because the goal was to evaluate a novel method in the context of unique research questions/objectives. While it may serve individual research goals, the summarized results are often difficult and time-consuming to re-produce because of missing details (e.g., parameter settings, pre-processing details) and non-deterministic nature of the algorithm (e.g., K-means).

The absence of assessment metrics at the dataset level means that researchers must repeat experiments in order to view the tradeoffs \cmmnt{across}among methods, thereby wasting precious resources and often delaying publications. The benchmark provided in this paper is intended to relax some of the burdens on researchers to foster more objective benchmark studies.

%% file: Methods.tex
\section{Benchmark Methods}\label{sec:benchmark_methods}

The benchmark methods comprise clustering methods (Section~\ref{sec:methods}) and evaluation methods (Section~\ref{sec:evaluation_framework}).

\subsection{Clustering methods}\label{sec:methods}
There are two major design criteria in clustering methods: the clustering algorithm and the distance measure. Eight clustering methods are used in this benchmark (see Table~\ref{tbl:algorithms}). They represent three categories of clustering algorithms --- partitional, density-based, and hierarchical --- and three distance measures --- Euclidean, dynamic time warping (DTW), and shape-based.
This subsection summarizes the clustering algorithms \cmmnt{(Section~\ref{sec:clustering_algs})} and distant measures.  

\begin{table}
\caption{Eight benchmark clustering methods.}\label{tbl:algorithms}
\centering
\begin{tabular}{|c|c|c|}
\hline
\multicolumn{2}{|c|}{\bf Clustering Method} & \multirow{2}{*}{\bf Category}\\
\cline{1-2}
{\bf Clustering algorithm}  & {\bf Distance measure} & \\
\hline
K-means  & Euclidean &  \multirow{5}{*}{Partitional}\\
K-medoids  &  Euclidean & \\
Fuzzy C-means &  Euclidean &\\
K-means  &  Shape-based &\\
K-means  &  Dynamic time warping &\\
\hline
Density Peaks   & Euclidean & \multirow{2}{*}{Density-based}\\
Density Peaks  &  Dynamic time warping &\\
\hline
Agglomerative  & Euclidean & Hierarchical\\
\hline
\end{tabular}
\end{table}

\subsubsection{Clustering algorithms}\label{sec:clustering_algs}
Choice of clustering algorithms may depend on the strategy used to maximize the intra-group similarity and minimize the inter-group similarity. The algorithms considered in this benchmark cover three popularly used categories of such strategies, each described below. 

\subsubsection*{Partitional}\label{sec:partitional}

\begin{figure*}[!htb]
\centering

\begin{subfigure}[b]{0.30\textwidth}
\centering
\includegraphics[width=1\textwidth]{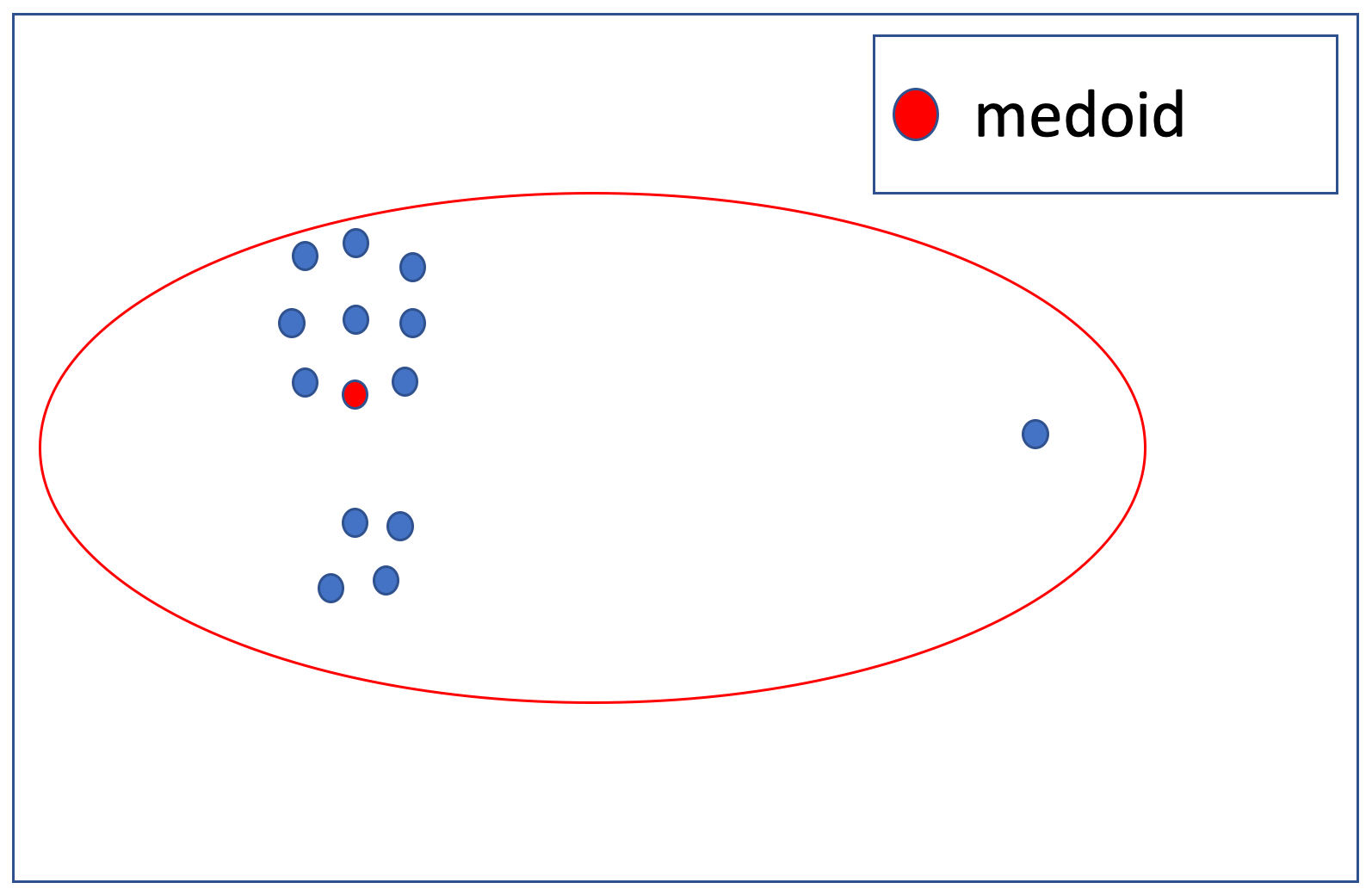}
\caption{}\label{fig:medoid}
\end{subfigure}
\begin{subfigure}[b]{0.30\textwidth}
\centering
\includegraphics[width=1\textwidth]{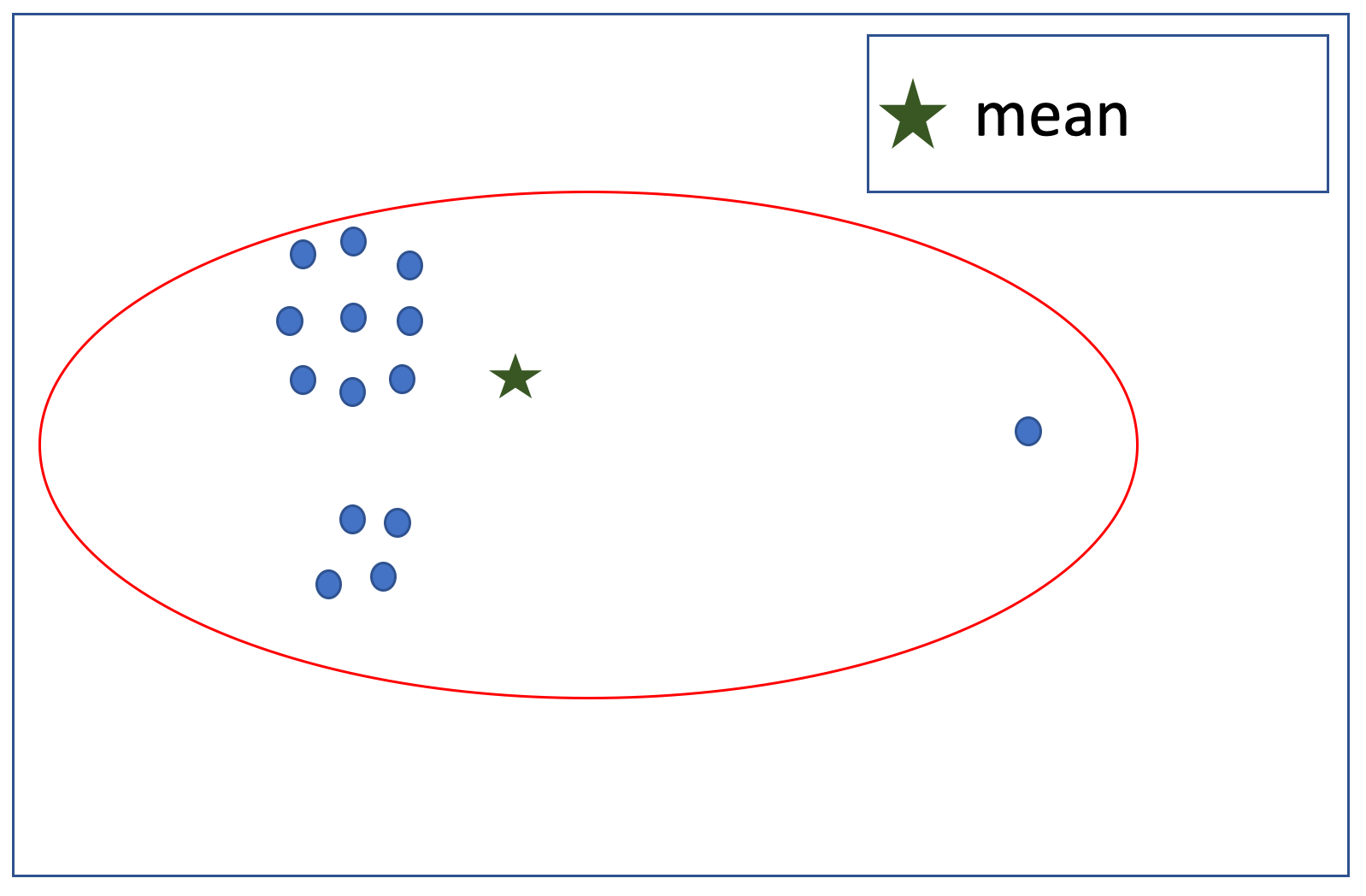}
\caption{}\label{fig:mean}
\end{subfigure}
\begin{subfigure}[b]{0.30\textwidth}
\centering
\includegraphics[width=1\textwidth]{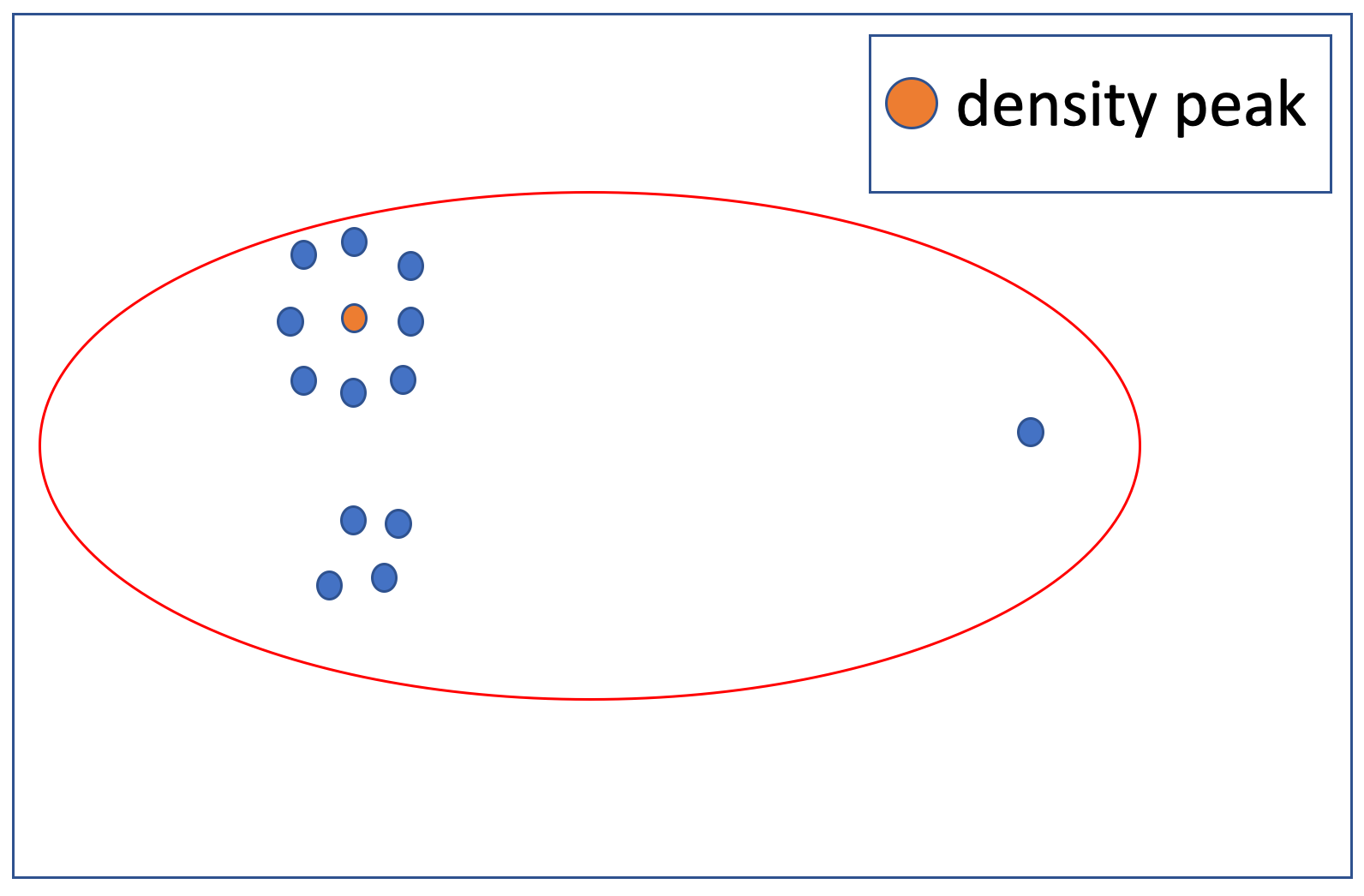}
\caption{}\label{fig:density}
\end{subfigure}
\caption{Different types of centroids: (a) medoid in K-medoids, (b) centroid in K-means, and (c) density peak in Density Peaks.}\label{fig:centers}
\end{figure*}
Three partitional clustering algorithms, K-means~\cite{macqueen1967}, K-medoids~\cite{Kaufman1990}, and Fuzzy C-means~\cite{Bezdek1981}, are selected based on their popularity~\cite{Ali2019} and known accuracy for time series data clustering~\cite{Paparrizos2017}. 
Note K-means with shape-based distance is K-shape~\cite{Paparrizos2017}. These partitional algorithms generate spherical clusters that are similar in size~\cite{WARREN2005}; and optimize clustering by minimizing the distance between each cluster center (a.k.a. centroid) and the data points within that cluster. A centroid may or may not be an actual data point, depending on the algorithm -- it is for K-medoids and not for K-means and Fuzzy C-means (see Figure~\ref{fig:medoid} and Figure~\ref{fig:mean}). 

All three of these partitional algorithms require that one input parameter be specified -- the number of clusters ($k$). Given $k$, the algorithm iterates over two phases: (1) calculate centroids, and (2) assign data points to their closest centroid, until some termination condition (e.g., number of iterations or convergence) is met. For all three algorithms used in this benchmark, the initial centroids are chosen at random, making the algorithm non-deterministic; all subsequent centroids are calculated so as to minimize the distance to all other data points within the given cluster. 

While K-means and K-medoids are hard clustering algorithms (i.e., producing non-overlapping partitions), Fuzzy C-means is a soft clustering algorithm (i.e., producing overlapping partitions). In this benchmark, the Fuzzy C-means clustering results are similar to that of a hard clustering algorithm, as each data point is assigned to the cluster that has the highest probability. 
There are several techniques for improving the clustering accuracy of these algorithms including---performing $z$-score normalization on the input~\cite{Mohamad2013}, or invoking the algorithm multiple times using different random seeds to select the clusters with the highest intra-cluster similarity and the lowest inter-cluster similarity. This benchmark excludes using such techniques, per restrictions~\ref{enum:no_preprocessing} and ~\ref{enum:established} (see Section~\ref{sec:Introduction}).

\subsubsection*{Density-based}\label{sec:density}
Density Peaks~\cite{densitypeaks} was selected as the representative for density-based algorithms due to its recent popularity, particularly for time series clustering~\cite{Nurjahan2016}. Unlike other density-based algorithms~\cite{Martin1996}, Density Peaks is not sensitive to the ``density parameter'' but needs the number of clusters, $k$, as one of the inputs.  This makes it a good fit for this benchmark, where $k$ is assumed to be known and no assumptions are made for other input parameters.

The Density Peaks algorithm generates cluster centroids (called ``density peaks'') that are surrounded by neighboring data points that have lower local density (see Figure~\ref{fig:density}) and are relatively farther from data points with a higher local density~\cite{densitypeaks}.
The algorithm has two phases. It first finds centroids (density peaks), and then assigns data points to the closest centroid. The algorithm requires two input parameters: the number of clusters ($k$) and the local neighborhood distance $d$ (wherein the local density of a data point is calculated). While the value of $k$ is assumed to be known in this benchmark, the value of $d$ is determined as the distance wherein the average number of neighbors is 1 to 2\% of the total number of observations in the dataset, following a rule of thumb proposed by the original authors~\cite{Rodriguez1492}.

\subsubsection*{Hierarchical}\label{sec:hierarchical}
\begin{figure}
\centering
\includegraphics[width=0.4\textwidth]{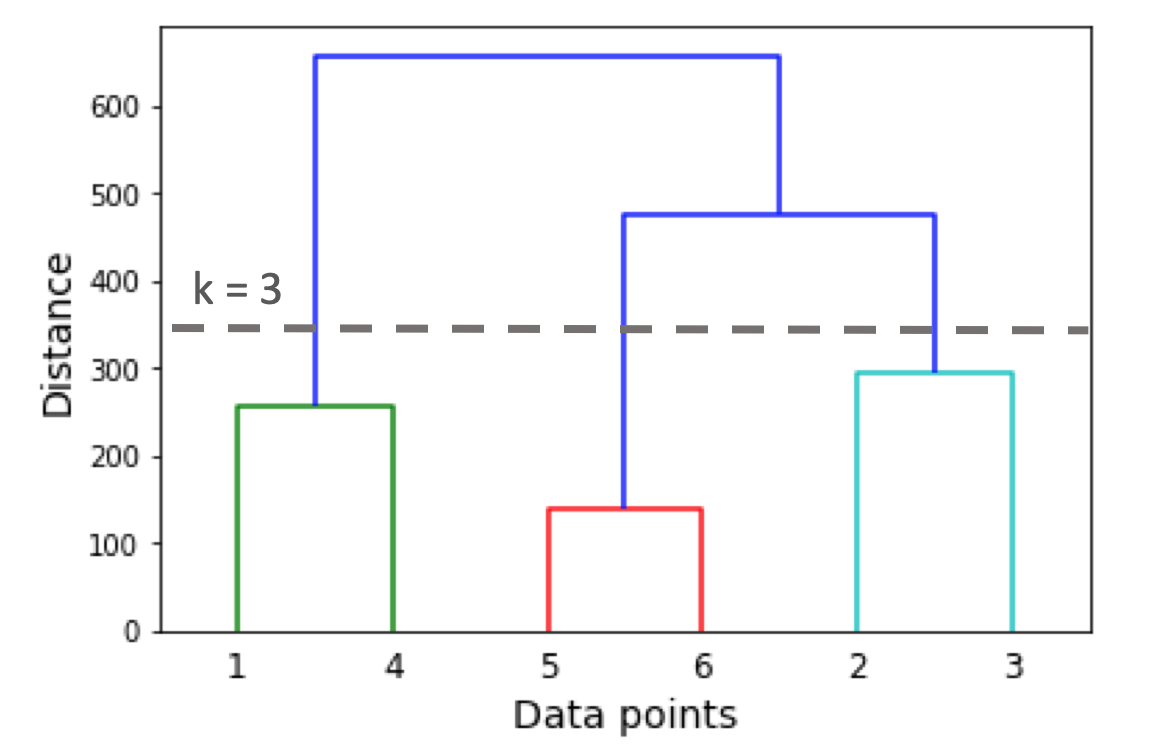}
\caption{Agglomerative clustering.}\label{fig:hie}
\end{figure}

A hierarchical clustering algorithm can be Agglomerative (bottom-up) or divisive (top-down). In the former, each data point begins as its own cluster and cluster pairs are merged as the algorithm moves up the hierarchy. In the latter, all data points are initially assigned to a single cluster and clusters are split as the algorithm moves down the hierarchy. Because of its popularity over divisive clustering~\cite{WARREN2005}, Agglomerative clustering is used in this benchmark. 

The algorithm has two phases. It first initializes each data point into its own cluster and then repeatedly merges the two nearest clusters into one until there are $k$ clusters (see Figure~\ref{fig:hie}). The value of $k$ is an input to the algorithm. There are several options for measuring the distance between pairs of clusters. Ward's linkage, which minimizes the variance of data points in the merged clusters~\cite{Anna2019}, is used in this benchmark due to its popularity and also its similarity to the optimization strategy of the partitional clustering methods. Other popular distance measures include single-linkage (minimum distance between a pair of data points belonging to different clusters) and complete-linkage (maximum distance between a pair of data points belonging to different clusters)~\cite{Li2017}.

\subsubsection{Distance measures}\label{sec:clustering_dists}
The choice of distance measure is the other criterion that has a direct impact on the clustering performance. This section discusses the three distance measures used in this benchmark.

\subsubsection*{Euclidean distance}\label{sec:Euclidean}
The most common distance measure used in a broad range of application is the Euclidean distance~\cite{Faloutsos1994}.
Equation~\ref{eq:euc} shows how the Euclidean distance  $d(T1, T2)$ is calculated between two time series $T1 = (T1_1,T1_2,...,T1_n)$ and $T2 = (T2_1, T2_2, ..., T2_n)$. 
\begin{equation}
d(T1,T2) = \sqrt{\sum_i^n (T1_i -T2_i)^2}\label{eq:euc}
\end{equation}

\subsubsection*{Dynamic time warping}\label{sec:DTW}
\begin{figure*}[!ht]
\centering
\includegraphics[width=0.8\textwidth]{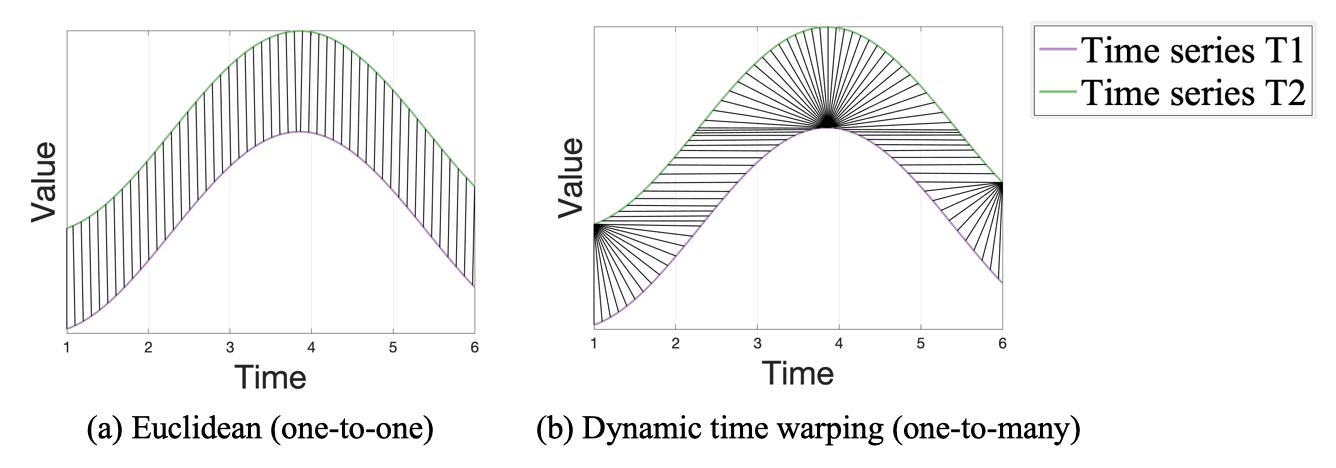}
\caption{Alignment between two times series for calculating distance.} 
\label{fig:DTW_EUC}
\end{figure*}

Dynamic time warping (DTW) is a mapping of points between a pair of time series, $T1$ and $T2$ (see Figure~\ref{fig:DTW_EUC}) designed to minimize the pairwise Euclidean distance. It is becoming recognized as one of the most accurate similarity measures for time series data~\cite{Paparrizos2017, Rakthanmanon2012, johnpaul2020}. The optimal mapping should adhere to three rules. 
\begin{itemize}
    \item Every point from $T1$ must be aligned with one or more points from $T2$, and vice versa.
    \item The first and last points of $T1$ and $T2$ must align. 
    \item No cross-alignment is allowed, that is, the warping path must increase monotonically.
\end{itemize}

DTW is often restricted to mapping points within a moving window. In general, the window size could be optimized using supervised learning with training data; this, however, is not possible with clustering as it is an unsupervised learning task. Paparrizos and Gravano~\cite{Paparrizos2016} found 4.5\% of the time series length to be the optimal window size when clustering 48 of the time series datasets in the UCR archive; as a result, we use a fixed window size of 5\% in this benchmark study. 

Density Peaks with DTW as the distance measure can be computationally infeasible for larger datasets because the Density Peaks algorithm is non-scalable of $O(n^2)$ complexity~\cite{Paparrizos2017}. We employ a novel pruning strategy (see TADPole~\cite{Nurjahan2016}) to speed up the algorithm by pruning unnecessary DTW distance calculations.

\subsubsection*{Shape-based distance}\label{sec:shape-based}

Shape-based distance is both shift-invariant and scale-invariant~\cite{Paparrizos2016}, that is, not affected by the shifting or scaling of the time series data. It calculates the cross-correlation between two time series and produces a distance value between 0.0 to 2.0, with 0.0 indicating that the time series are identical and 2.0 indicating maximally different shapes. To ensure the distance measure is scale-invariant, the original time series, $T$, is z-normalized to $T'$ as follows~\cite{Paparrizos2016}:
\begin{equation}\label{eq:zscore}
    T' = \frac{T - \mu}{\sigma}
\end{equation}
\noindent so $T'$ has mean $\mu'= 0$ and standard deviation $\sigma' = 1$.

\subsection{Evaluation methods}\label{sec:evaluation_framework}

The purpose of this benchmark study is to assess the performance of the eight clustering algorithms on the 112 datasets, as well as the impact of changing design choices in either clustering algorithms or distance measures. To this end, the evaluation framework and select assessment metrics \cmmnt{(Section~\ref{sec:assessment} and Section~\ref{sec:spread})} \cmmnt{(Section~\ref{sec:eval_methodology})} are discussed in this section.

\subsubsection{Assessment metrics}\label{sec:assessment}
\begin{figure}
\centering
\begin{subfigure}[b]{0.3\textwidth}
\centering
\includegraphics[width=1\textwidth]{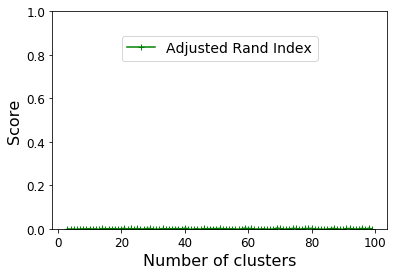}
\caption{}\label{fig:ARI}
\end{subfigure}
\quad%
\begin{subfigure}[b]{0.3\textwidth}
\centering
\includegraphics[width=1\textwidth]{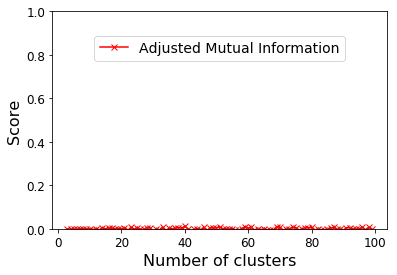}
\caption{}\label{fig:Mutual}
\end{subfigure}
\quad%
\begin{subfigure}[b]{0.3\textwidth}
\centering
\includegraphics[width=1\textwidth]{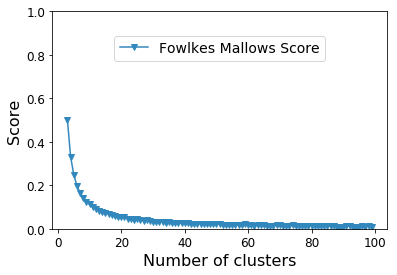}
\caption{}\label{fig:Fowlkes}
\end{subfigure}
\medskip\\
\begin{subfigure}[b]{0.3\textwidth}
\centering
\includegraphics[width=1\textwidth]{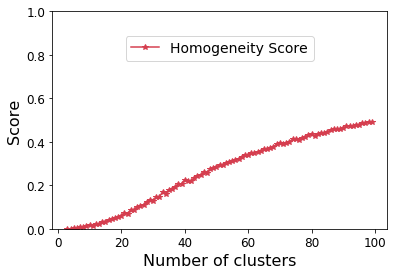}
\caption{}\label{fig:Homogeneity}
\end{subfigure}
\quad%
\begin{subfigure}[b]{0.3\textwidth}
\centering
\includegraphics[width=1\textwidth]{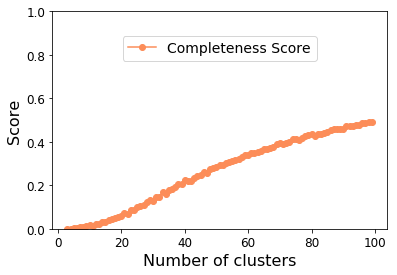}
\caption{}\label{fig:Completeness}
\end{subfigure}
\quad%
\begin{subfigure}[b]{0.3\textwidth}
\centering
\includegraphics[width=1\textwidth]{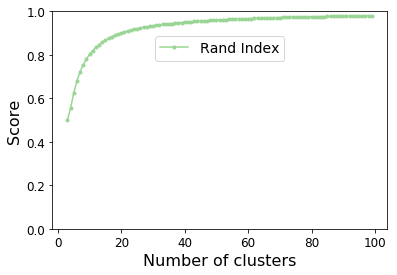}
\caption{}\label{fig:RandIndex}
\end{subfigure}
\caption{Accuracy scores resulting from randomly assigning 1000 data points to a varying number of clusters.}
\label{fig:measures}
\end{figure}
%
Metrics for assessing clustering output may be external or internal. External measures are used when the class labels are available for individual data points. Examples include the Rand Index~\cite{hubert1985}, Adjusted Rand Index (ARI)~\cite{Santos2009}, Adjusted Mutual Information~\cite{Romano2016}, Fowlkes Mallows index~\cite{Fowlkes1983},  Homogeneity~\cite{Rosenberg2017}, and Completeness~\cite{Rosenberg2017}.  Internal measures quantify the goodness of clusters based on a optimization objective for the clustering output, without the need for class labels; examples include Silhouette score~\cite{Rousseeuw1987}, Davies-Bouldin index~\cite{Davies1979}, Calinski- Harabasz index~\cite{Cali1974}, the I-index~\cite{Maulik2002} and sum of square errors (SSE).

We used all the external measures listed above in this benchmark because having the class labels provided in the UCR archive makes the evaluation independent of the algorithms optimization function. 
Despite the popularity of the Rand Index (Figure~\ref{fig:RandIndex}) for prior 
UCR archive studies~\cite{Paparrizos2016,Paparrizos2017,Nurjahan2016}, we find the adjusted measures more suitable for clustering because they are independent of the number of clusters. As demonstrated in Figure~\ref{fig:measures}, the accuracy scores resulting from random cluster assignment are consistently low as the number of clusters varies for the two adjusted measures (Figures~\ref{fig:ARI} and \ref{fig:Mutual}), while this is not the case for the other measures. In this work, the Adjusted Rand Index was selected as the default measure.

For the partitional algorithms in this benchmark, all of which are non-deterministic, the scores reported for each external measure are the average over ten runs using randomly selected initial centroids.

\subsubsection*{Adjusted Rand Index}\label{sec:randindex}

The Adjusted Rand Index is the adjusted-for-chance version of the more commonly used Rand Index. Given two sets of clusters, $X$ and $Y$, and a contingency table where each cell $n_{ij}$ is the number of elements in both the $i^{th}$ cluster of X and the $j^{th}$ cluster of Y, the Adjusted Rand Index is calculated as shown in Equation~\ref{eq:ari}.

\begin{equation}
\text{Adjusted Rand Index} =
    \frac{ 
       \sum_i^j \binom{n_{ij}}{2} - 
        [ \sum_i \binom{a_i}{2} \sum_j \binom{b_j}{2} ] / \binom{n}{2}
    }
    { 
        \frac{1}{2} [ \sum_i \binom{a_i}{2} + \sum_j \binom{b_j}{2} ]
        - [ \sum_i \binom{a_i}{2} \sum_j \binom{b_j}{2} ] / \binom{n}{2}
    }\label{eq:ari}
\end{equation}
\noindent where $a_i$ is the sum of the $i^{th}$ row and $b_{j}$ is the sum of the $j^{th}$ column in the contingency table.

\subsubsection*{Spread between clustering outputs}\label{sec:spread}

The measure of spread is used to quantify how much the accuracy of the two clustering methods differ from each other over multiple datasets (see Equation~\ref{eq:spread}).

\begin{equation}\label{eq:spread}
\text{Spread} = \frac{\sum_{i=1}^{n}{(A1_i - A2_i)^2}}{n}
\end{equation}
\noindent where $A1_i$ and $A2_i$ are the accuracy scores of the two methods for dataset $i$; and n is the total number of datasets. 
%

\subsubsection{Evaluation framework}\label{sec:eval_methodology}

Researchers will often design an evaluation framework for assessing accuracy because what constitutes ``good'' with respect to the assessment metrics may vary depending on the research question.
%
One of the simplest approaches is to rank the performance of each clustering method and tally the number of winning performances across all available (in this work 112) datasets. This approach, however, is not without bias, as it depends on the distribution of both the datasets and clustering methods. For instance, in this work there are five partitional methods and one density-based method. If one half the datasets are amenable to partitional and the other half to density-based, this evaluation metric will bias the  density-based method because the tally for the partitional methods would be partitioned across the five datasets.
On the other extreme, if pairwise comparison were performed on all clustering methods, it would result in 28 ($=\binom{8}{2}$) pairwise comparisons for each of the 112 datasets (i.e., 3,136 comparisons). More importantly, a pairwise comparison assumes that every algorithm is designed to achieve the same result.

Based on the above challenges, we designed a phased evaluation approach in this benchmark study. This approach first compares the eight clustering methods, and then controls for either the distance measure or clustering algorithm while evaluating the impact of changing the other.

\paragraph{Phase 1} All eight methods are compared using all datasets, and the resulting accuracy is averaged over all datasets for each method. 
\paragraph{Phase 2} Partitional algorithms with Euclidean distance are compared to select the one that achieves the highest accuracy on the largest number of datasets.
\paragraph{Phase 3} Different distance measures are compared using the partitional algorithm selected in Phase 2. 
\paragraph{Phase 4} Clustering algorithms belonging to different categories are compared using Euclidean distance. Among them, the partitional algorithm is the one selected in Phase 2 (i.e., K-means with Euclidean distance).
\paragraph{Phase 5} Density Peaks algorithm using Euclidean distance is compared with Density Peaks algorithm using DTW. %
\paragraph{Phase 6} Density Peaks algorithm using DTW \cmmnt{(TADPole)} is compared with the partitional algorithm selected in Phase 2 but using DTW. 
\smallskip

In Phase 1, we report the average ARI and standard deviation across all datasets. In each subsequent phase, we report the number of datasets (called ``winning count'') for which an algorithm or a distance measure achieved the highest ARI, and refine the comparison with the measure of spread (see Section~\ref{sec:spread}) and the associated scatter plots. Here, datasets that result in an ARI score lower than 0.05 are excluded from winning counts since scores that approach 0.00 represent random assignment.

%% file: Results.tex
\section{Benchmark Test Results}\label{sec:results}
This section provides the results of dataset-level assessment (Section~\ref{sec:dataset-level}) and the phased evaluation (Section~\ref{sec:phased_evaluation}), and discusses the results (Section~\ref{sec:discussion}). 

\subsection{Dataset-level assessment}\label{sec:dataset-level}

Appendix~\ref{apx:raw_ARI}  shows the Adjusted Rand Index (ARI) scores for all eight clustering methods on the 112 short-listed datasets (see Section~\ref{sec:Introduction}) in the UCR archive (\cmmnt{Tables~\ref{tbl:ari_scores_1} to \ref{tbl:ari_scores_3}}Table~\ref{tbl:ari_scores}), and the spread of ARI scores (Table~\ref{tbl:spreadscores}) between each pair of clustering methods. Additionally, in line with  the restriction~\ref{enum:gp_raw} (dtatset-level assessment; see Section~\ref{sec:Introduction}), the \cmmnt{raw} scores of each clustering method on each dataset tested for all the six external measures (see Section~\ref{sec:assessment}) are available at GitHub~\cite{clusteringBenchmark} along with the source codes.

\subsection{Phased evaluation}\label{sec:phased_evaluation}

\begin{figure}
\centering
\includegraphics[width=0.55\textwidth]{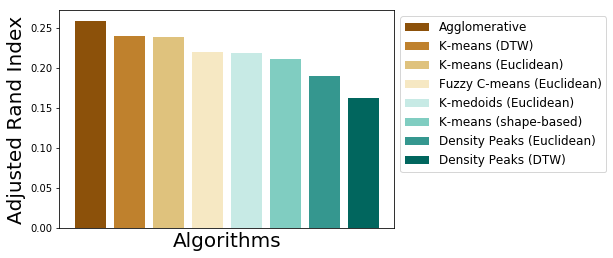}
\captionof{figure}{Average ARI for each clustering method in Phase 1.}\label{fig:phase1}

\captionof{table}{Average ARI and standard deviation for each clustering method in Phase 1.}\label{tbl:phase1}
\begin{tabular}{|c|c|c|c|c|}
\hline
\multicolumn{2}{|c|}{\bf Clustering Method} & \multirow{2}{*}{\bf Category} & \multirow{2}{*}{\bf Avg ARI} & \multirow{2}{*}{\bf Stddev ARI}  \\
\cline{1-2}
{\bf Algorithm}& {\bf Distance measure} &   &  & \\
\hline
Agglomerative  &  Euclidean & Hierarchical & 0.26 & 0.26\\
\hline
K-means  & DTW & \multirow{5}{*}{Partitional} & 0.24 & 0.24\\
K-means  &  Euclidean & &0.24 & 0.24\\
Fuzzy C-means  &   Euclidean & &0.22 & 0.25 \\
K-medoids  &   Euclidean & &0.22 & 0.23\\
K-means  &   Shape-based & &0.21 & 0.22\\
\hline
Density Peaks  &  Euclidean & \multirow{2}{*}{Density-based} & 0.19 & 0.24 \\
Density Peaks \cmmnt{TADPole}  &  DTW & &0.16& 0.25 \\
\hline
\end{tabular}
\end{figure}

\paragraph{Phase 1 - Ranked comparison of all methods using average ARI}
Figure~\ref{fig:phase1} shows the average ARI's for each of the eight clustering methods in decreasing order; and Table~\ref{tbl:phase1} provides corresponding detail including the standard deviation of the ARI. The highest average ARI was 0.26 for the Agglomerative clustering using Ward linkage and Euclidean as distance measure; and the lowest average ARI was 0.16 for Density Peaks using DTW as distance measure. The standard deviations show wide variation in method performance.
%

\begin{figure}
\centering
\begin{subfigure}[b]{0.30\textwidth}
\centering
\includegraphics[width=1\textwidth]{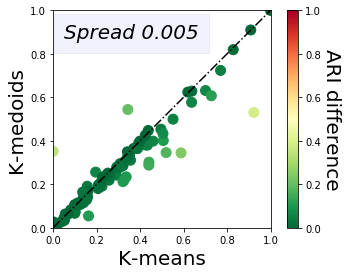}
\caption{K-medoids vs. K-means.}\label{fig:phase2a}
\end{subfigure}\qquad
\begin{subfigure}[b]{0.30\textwidth}
\centering
\includegraphics[width=1\textwidth]{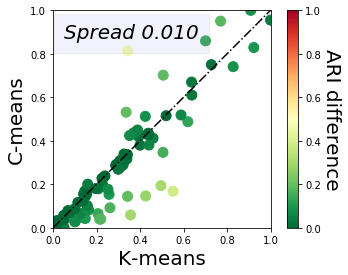}
\caption{Fuzzy C-means vs. K-means}\label{fig:phase2c}
\end{subfigure}\qquad
\begin{subfigure}[b]{0.30\textwidth}
\centering
\includegraphics[width=1\textwidth]{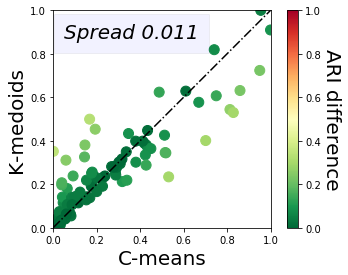}
\caption{K-medoids vs. Fuzzy C-means.}\label{fig:phase2b}
\end{subfigure}
\caption{Spread of ARI scores between each pair of the three clustering algorithms with Euclidean distance in Phase 2.}\label{fig:phase2}
\captionof{table}{Clustering algorithms with Euclidean distance in Phase 2.}
\begin{tabular}{|c|c|c|}
\hline
{\bf Algorithm}  & {\bf Winning count}\\
\hline
\multicolumn{2}{|c|}{\it Triple-wise}\\
\hline
K-means  & {54} \\
Fuzzy C-means  & {31} \\
K-medoids  & {18}  \\
\hline
\multicolumn{2}{|c|}{\it Pairwise}\\
\hline
K-means & 64\\
K-medoids & 17\\
\hline
K-means &54 \\ 
Fuzzy C-means & 27 \\
\hline
Fuzzy C-means & 41\\
K-medoids & 39\\
\hline
\end{tabular}\label{tbl:phase2}
\end{figure}

\paragraph{Phase 2 - Comparison of partitional algorithms using Euclidean distance}
Of the partitional clustering methods that use a Euclidean distance measure, K-means had a winning count of 54 datasets, while Fuzzy C-means and K-medoids performed best on 31 and 18 datasets, respectively, (see Table~\ref{tbl:phase2}). While K-means had a higher ARI score in almost twice as many datasets, differences in score values were minor, with a spread of only 0.005 against K-medoids (see Figure~\ref{fig:phase2a}) and only slightly larger (0.010) against Fuzzy C-means (see Figure~\ref{fig:phase2c}). 
This result is not surprising, given the similarity of methodology (all partitional using Euclidean distance) across the three algorithms.

\begin{figure}
\centering
\begin{subfigure}[b]{0.30\textwidth}
\centering
\caption{}\label{fig:phase3a}
\includegraphics[width=1\textwidth]{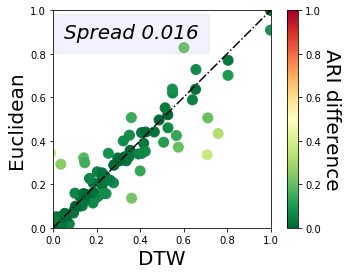}
\end{subfigure}\qquad
\begin{subfigure}[b]{0.30\textwidth}
\centering
\caption{}\label{fig:phase3b}
\includegraphics[width=1\textwidth]{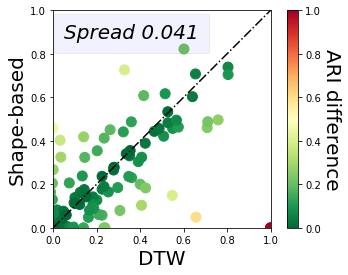}
\end{subfigure}\qquad
\begin{subfigure}[b]{0.30\textwidth}
\centering
\caption{}\label{fig:phase3c}
\includegraphics[width=1\textwidth]{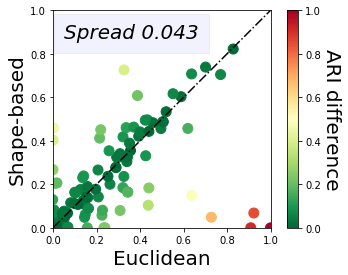}
\end{subfigure}
\caption{Spread of ARI scores between each pair of distance measures in Phase 3}\label{fig:phase3}
\captionof{table}{Different distance measures for K-means (from Phase 2) in Phase 3.}\label{tbl:phase3}
\begin{tabular}{|c|c|}
\hline
{\bf Distance measure} & {\bf Winning count}\\
\hline
\multicolumn{2}{|c|}{\it Triple-wise}\\
\hline
DTW & {32}\\
Shape-based & {31}\\
Euclidean & {28}\\
\hline
\multicolumn{2}{|c|}{\it Pairwise}\\
\hline
DTW &  45\\
Euclidean  &38\\
\hline
DTW & 52 \\
Shape-based& 38 \\
\hline
Shape-based & 45\\
Euclidean  & 44\\
\hline
\end{tabular}
\end{figure}

\paragraph{Phase 3 - Comparison of distance measures using selected partitional algorithm}
When we examine the winning counts for K-means (i.e., method that performed best in Phase 2) using the three distance measures, the tallies are 32, 31 and 28 for DTW, shape-based, and Euclidean, respectively (see Table~\ref{tbl:phase3}). A pairwise comparison between the distance measures also shows the wining counts to be 45 vs.~38 between DTW and Euclidean, 52 vs.~38 between DTW and shape-based, and 45 vs.~44 between shape-based and Euclidean. The scatter plots in Figure~\ref{fig:phase3} show the spreads between each of the paired distance measures. The shape-based distance has a relatively larger spread with each of the other two measures.   
As a side note, when the optimal DTW window size is assumed to be known, then it is trivial to understand that DTW will always achieve a score that is higher or equal to that of Euclidean distance, since the two measures are equivalent when the window size is 0.

\begin{figure}
\centering
\begin{subfigure}[t]{0.30\textwidth}
\centering
\includegraphics[width=1\textwidth]{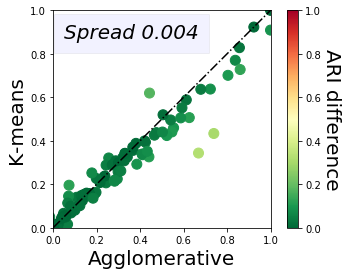}
\caption{K-means vs. Agglomerative.}\label{fig:phase4a}
\end{subfigure}
\quad
\begin{subfigure}[t]{0.30\textwidth}
\centering
\includegraphics[width=1\textwidth]{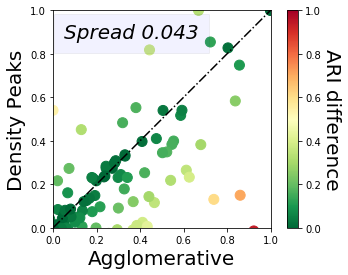}
\caption{Density Peaks vs. Agglomerative.}\label{fig:phase4b}
\end{subfigure}
\quad
\begin{subfigure}[t]{0.30\textwidth}
\centering
\includegraphics[width=1\textwidth]{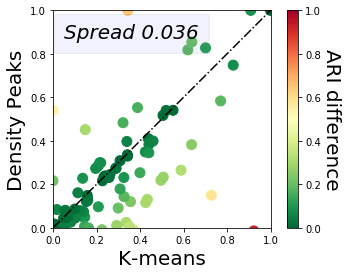}
\caption{Density Peaks vs. K-means.}\label{fig:phase4c}
\end{subfigure}
\caption{Different algorithms with Euclidean distance measure in Phase 4.}\label{fig:phase4}
\captionof{table}{Different algorithms with Euclidean distance measure in Phase 4.}\label{tbl:phase4}
\begin{tabular}{|c|c|}
\hline
{\bf Algorithm}  &  {\bf Winning count}\\
\hline
\multicolumn{2}{|c|}{\it Triple-wise}\\
\hline
Agglomerative  & {45} \\
K-means  & {21} \\
Density Peaks  & {19}  \\
\hline
\multicolumn{2}{|c|}{\it Pairwise}\\
\hline
Agglomerative & 57\\
Density Peaks & 26\\
\hline
Agglomerative & 52\\
K-means & 30\\
\hline
K-means & 60 \\
Density Peaks & 23\\ 
\hline
\end{tabular}
\end{figure}

\paragraph{Phase 4 - Comparison of clustering algorithms using Euclidean distance}
When we hold the distance measure (in this case, Euclidean distance)  constant and examine the winning counts across the clustering algorithms that use this distance measure, the tallies are 45, 21, and 19 in the order of Agglomerative, K-means, and Density Peaks. A pairwise comparison is also shown in Table~\ref{tbl:phase4}, where the winning counts are 57 vs.~26 between Agglomerative and Density Peaks, 52 vs.~30 between Agglomerative and K-means, and 60 vs.~23 between K-means and Density Peaks. Despite the difference in winning counts, the spreads of ARI values between Agglomerative and K-means (see Figure~\ref{fig:phase4a}) is fairly small compared with the spread of either method with Density Peaks (see Figure~\ref{fig:phase4b} and Figure~\ref{fig:phase4c}).

\begin{figure}
\centering
\includegraphics[width=0.3\textwidth]{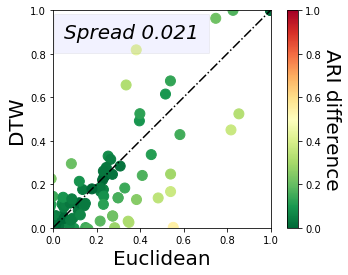}
\caption{Euclidean vs.~DTW for Density Peaks algorithm in Phase 5.}\label{fig:phase5}
\captionof{table}{Euclidean vs.~DTW for Density Peaks algorithm in Phase 5.}
\begin{tabular}{|c|c|}
\hline
{\bf Distance measure} & {\bf Winning count}\\
\hline
Euclidean & {45} \\
DTW & {31}\\
\hline
\end{tabular}
\label{tbl:phase5}
\end{figure}

\paragraph{Phase 5 - Comparison of Euclidean distance and DTW in Density Peaks algorithm}
The Density Peaks algorithm achieved a higher winning count (i.e., across 45 datasets; see Table~\ref{tbl:phase5}) when Euclidean distance was used as the distance measure compared to a count of 31 with DTW. Figure~\ref{fig:phase5} shows the spread of ARI scores between Euclidean distance and DTW to be 0.021.

\subsubsection*{Phase 6 - Comparison of Density Peaks and selected partitional algorithm using DTW}
\begin{figure}
\centering
\includegraphics[width=0.30\textwidth]{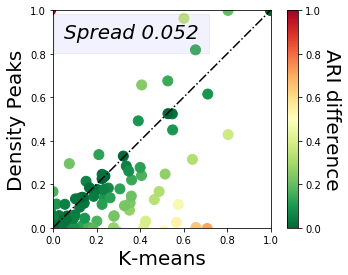}
\caption{DTW in Density Peaks and K-means (selected in Phase 2) in Phase 6.}\label{fig:phase6}
\captionof{table}{DTW in Density Peaks and K-means (selected in Phase 2) in Phase 6.}\label{tbl:phase6}
\begin{tabular}{|c|c|}
\hline
{\bf Algorithm} &  {\bf Winning count}\\
\hline
K-means  &   {60}\\
Density Peaks\cmmnt{TADPole} &  {24} \\
\hline
\end{tabular}
\end{figure}

Lastly, when the DTW distance measure is held constant, we may compare across the clustering algorithms that use this distance measure - Density Peaks 
and K-means. 
K-means achieved a higher winning count (i.e., winner across 60 datasets; see Table~\ref{tbl:phase6}) compared to a winning count of 24 for Density Peaks. But while the winning count appears positively skewed in favor of K-means, there are still a considerable number of datasets for which Density Peaks achieved higher ARI, and the spread of ARI scores (see Figure~\ref{fig:phase6}) was the largest (0.052) observed in the six phases.

%% file: Discussion.tex
\subsection{Discussion}\label{sec:discussion}

This section analyzes the results of each evaluation phase and provides concluding remarks summarizing the analysis. 

\paragraph{Phase 1  - Ranked comparison of all methods using average ARI}
The high standard deviations associated with the average ARI of Table~\ref{tbl:phase1} suggest that accuracy is dependent on which clustering method is used on which dataset; and that it may be fair to conclude that we have no clear winner in this benchmark.
This high variability also suggests that using a simple winning count of dataset-level assessment as the only means of evaluation, may be very misleading. While reporting counts of win-lose-tie for clustering method accuracy has become common practice in the literature, the UCR archive authors describe it as not that useful~\cite{UCRArchive2018}. In light of these issues, we used both winning counts and the ARI scores in this benchmark and reinforced the measures with ARI score scatter plots and the associated spreads. 

\paragraph{Phase 2 - Comparison of partitional algorithms using Euclidean distance}
When comparing the three partitional algorithms that use the Euclidean distance measure, a researcher may well select K-means based on the winning count (see Table~\ref{tbl:phase2}), especially without adequate prior knowledge of how the algorithm performs on the individual datasets. However, the selection may likely change when the user has knowledge of the dataset and/or application at hand. For instance, K-medoids is more resilient to outliers, because the medoids are not as sensitive to the presence of outliers as say, the centroids in K-means. 
In another example, Fuzzy C-means may be preferred over K-means given a dataset where the membership of data points are ``soft'', 
as in the case when categorical classes have numerical attribute values that overlap.
As an aside, Fuzzy C-means shows a larger spread of ARI scores against K-means (Figure~\ref{fig:phase2c}) and K-medoids (Figure~\ref{fig:phase2b}), indicating that changing from K-means to the fuzzy mechanism of C-means has more impact on the final clustering than changing from means to medoids. 

\paragraph{Phase 3 - Comparison of distance measures using selected partitional algorithm}
The results in Table~\ref{tbl:phase3} appear to suggest that the winning count does not favor the shape-based distance measure in the same manner that it did in a prior study~\cite{Paparrizos2017} that used 85 datasets in the UCR archive compared to the 112 datasets (and different evaluation criteria) used in this benchmark study. 
The larger spreads observed when one distance measure is shape-based (Figures~\ref{fig:phase3b} and \ref{fig:phase3c}) suggest the method is useful as the best distance measure for a nontrivial number of datasets, and therefore, should be considered in a pool of potential clustering methods. We believe the larger spread may be a result of the shape-based distance measure's 
lack of sensitivity to the magnitudes and shifts in time series data compared with the Euclidean measure, or for that matter, DTW (for which the underlying distance measure is also Euclidean), which therefore results in a different partitioning.

\paragraph{Phase 4 - Comparison of clustering algorithms using Euclidean distance}
The very small spread in Figure~\ref{fig:phase4a} shows similar performance for the K-means and Agglomerative algorithms on most datasets in the archive. With Agglomerative clustering, this can be attributed to the use of Ward's linkage, which merges the two clusters that when combined provide the minimum increase in variance. This optimization using Ward's linkage has some similarity to optimizing the centroids in K-means (i.e., minimizing the total variance within cluster). Using a different linkage criteria such as ``complete'' linkage does not bias clusters to be as spherical as Ward linkage (and for that matter K-means). Such a change will result in different clusters when compared to K-means. Specifically, with complete linkage, Agglomerative clustering has a measure of spread of 0.026 when compared to K-means, and an average ARI of $0.17 \pm 0.24$.

\paragraph{Phase 5 - Comparison of Euclidean distance and DTW in Density Peaks algorithm}
The spread (0.021) between DTW and Euclidean (see Figure~\ref{fig:phase5}) in Density Peaks algorithm is relatively consistent with spread (0.016) between DTW and Euclidean in K-means algorithm (see Figure~\ref{fig:phase3a}). These medium to high level of spread values indicate the difference of clusters formed when using DTW as opposed to Euclidean distance.
Density Peaks is an $O(n^2)$ complexity algorithm (where $n$ is the number of data points) that when used with DTW may become computationally infeasible for large datasets. The TADPole method~\cite{Nurjahan2016}, with its novel pruning strategy, makes Density Peaks with DTW feasible enough for use on large datasets in the archive. However, even with this accelerated TADPole, the largest 20 datasets of the archive took 32 days to cluster on a dual 20-Core Intel Xeon E5-2698 v4 2.2 GHz machine with 512 GB 2,133 MHz DDR4 RDIMM.

\paragraph{Phase 6 - Comparison of Density Peaks and selected partitional algorithm using DTW}
When using DTW as a distance metric, K-means and Density Peaks produce different clusters as indicated by the relatively higher spreads of ARI 0.052 (see Figure~\ref{fig:phase6}), which is consistent with the somewhat high spread 0.036 observed between the two methods (see Phase 4 with Euclidean distance, Figure~\ref{fig:phase4c}). This result is counter-intuitive given that both K-means and Density Peaks form spherical clusters by assigning data points to the closest centroid, and leads one to speculate that the cause may be the fundamentally different locations of the centroids in the K-means and Density Peaks algorithms (see Figure~\ref{fig:centers}).

\paragraph{Concluding remarks}

Overall, this benchmark study shows that among all methods tested, the variation in performance, as measured by the average and standard deviation of ARI (see Table~\ref{tbl:phase1} and Figure~\ref{fig:phase1}), is higher than the variation observed across winning counts (Table~\ref{tbl:phase2} to Table~\ref{tbl:phase6}). Notably, there is no one method that performs better than the others for all datasets in this benchmark, and that method performance is dependent on the datasets, as well as the evaluation criteria (i.e., objective). 
Similar findings for time series representation methods and distance measures were made in an earlier benchmark study using UCR archive~\cite{Hui2010}.
In light of these findings, as well as noting that exploratory cluster analysis typically involves multiple clustering methods rather than a single method to identify clusters of interest, cluster analysis should be conducted by selecting a pool of methods that produce different clusters, rather than those that produce relatively similar clusters. In other words, select methods that show greater spread (i.e., combination of average accuracy scores and their spread) than those with higher winning counts. 
Methods with higher spreads of ARI are likely to provide different clusters for the same dataset---all of which may be valid depending on the target research goal. For instance, using three algorithms with higher spread values (e.g., Density Peaks (DTW), K-means (shape-based), and K-means (Euclidean) of Figure~\ref{fig:phase3c}, Figure~\ref{fig:phase4c} and Figure~\ref{fig:phase6}) on the same dataset are more likely to provide three dissimilar clustering outputs compared to those generated using K-means (Euclidean), K-medoids (Euclidean), and Fuzzy C-means (Euclidean) (lower spread values in Figure~\ref{fig:phase2}).

%% file: Conclusion.tex
\section{Conclusion}\label{sec:conclusion}
This paper reports benchmark test from applying eight popular time series clustering methods on 112 datasets in the UCR archive. One essential goal of the benchmark is to make the results available and reusable to other researchers. In this work, we laid out six restrictions to help reduce bias. Eight popular clustering methods were selected to cover three categories of clustering algorithms (i.e., partitional, density-based, and hierarchical) and three distance measures (i.e., Euclidean, Dynamic time warping, and shape-based). The dataset-level assessment metrics are reported using six external evaluation measures. Adjusted Rand Index was selected as the default measure for discussion in this paper. A phased evaluation framework was designed such that in each phase only one of the two building blocks of a clustering method---algorithm and distance measure---is varied at a time. Benchmark results show the overall performance of the eight algorithms to be similar with high variance across different datasets. Discussion of the results helps highlight the importance of creating a pool of clustering methods with high spread in accuracy scores for effective exploratory analysis.

%% file: Acknowledgements.tex
\section*{Acknowledgements}
This project was supported by the grant from the Barrett Foundation and Gund Institute for Environment through a Gund Barrett PhD Fellowship. This material is based upon work partially supported by the National Science Foundation under VT EPSCoR Grant No. NSF OIA 1556770. We thank Drs. Patrick J. Clemins and Scott Hamshaw, Research Assistant Professors at the University of Vermont, for support in using Vermont EPSCoR's high performance computing resources. We also thank Dr. Eamon Keogh for his invaluable feedback, and all other curators and administrators of the UCR archive without which this work would not have been possible.


%% file: Appendix.tex
\appendix

\section{Dataset-Level Assessment Results}\label{apx:raw_ARI}
\input{results_table.tex}

\input{spread_table.tex}

%% file: results_table.tex
\begin{small}
\begin{center}
\begin{longtable}{|p{5.8cm}|p{0.9cm}|p{0.9cm}|p{0.9cm}|p{0.9cm}|p{0.9cm}|p{0.9cm}|p{0.9cm}|p{0.9cm}|}
\caption{\normalsize ARI scores of the eight clustering methods on the 112 datasets in the UCR archive.}\label{tbl:ari_scores}\\
\hline
Dataset name & K-mean-Euc & K-med-Euc & K-mean-shape & K-mean-DTW & C-mean-Euc & D-Peaks-Euc & D-Peaks-DTW & Agglo-Euc\\
\hline
\endfirsthead
 %
\multicolumn{9}{c}{{\tablename\ \thetable{} -- continued from previous page}}\\
\hline
Dataset name & K-mean-Euc & K-med-Euc & K-mean-shape & K-mean-DTW & C-mean-Euc & D-Peaks-Euc & D-Peaks-DTW & Agglo-Euc\\
\hline
\endhead
\hline
\multicolumn{9}{|r|}{{Continued on next page}}\\
\hline
\endfoot
\hline
\endlastfoot
ACSF1&0.16&0.17&0.14&0.10&0.20&0.13&0.06&0.15\\
Adiac&0.25&0.25&0.24&0.23&0.18&0.23&0.11&0.18\\
ArrowHead&0.20&0.26&0.18&0.23&0.18&0.27&0.25&0.07\\
Beef&0.15&0.14&0.11&0.12&0.17&0.05&0.09&0.07\\
BeetleFly&0.05&0.04&0.04&0.01&0.00&0.04&0.11&-0.02\\
BirdChicken&0.04&0.03&0.07&0.00&0.04&0.00&0.05&0.04\\
BME&0.14&0.16&0.23&0.36&0.12&0.23&0.22&0.18\\
Car&0.14&0.14&0.13&0.20&0.16&0.05&0.03&0.11\\
CBF&0.33&0.22&0.73&0.33&0.34&0.14&0.10&0.44\\
Chinatown&0.16&0.19&-0.05&0.24&0.18&-0.07&-0.08&0.16\\
ChlorineConcentration&0.00&0.00&0.00&0.00&0.00&0.00&0.00&0.00\\
CinCECGTorso&0.15&0.14&0.06&0.21&0.04&0.45&0.34&0.13\\
Coffee&0.34&0.54&0.16&-0.01&0.81&1.00&1.00&0.67\\
Computers&0.00&0.00&0.07&0.00&0.00&0.00&0.01&0.00\\
CricketX&0.10&0.07&0.16&0.13&0.03&0.04&0.14&0.11\\
CricketY&0.13&0.11&0.18&0.14&0.07&0.08&0.11&0.14\\
CricketZ&0.10&0.07&0.16&0.13&0.03&0.05&0.14&0.12\\
Crop&0.31&0.28&0.08&0.31&0.28&0.18&0.18&0.33\\
DiatomSizeReduction&0.83&0.82&0.82&0.60&0.74&0.75&0.96&0.86\\
DistalPhalanxOutlineAgeGroup&0.39&0.39&0.42&0.51&0.42&-0.04&-0.02&0.42\\
DistalPhalanxOutlineCorrect&0.00&0.00&0.00&0.00&0.00&0.00&-0.02&0.00\\
DistalPhalanxTW&0.43&0.38&0.50&0.76&0.43&0.13&-0.05&0.74\\
DodgerLoopDay&0.23&0.23&0.08&0.17&0.20&0.22&0.18&0.20\\
DodgerLoopGame&0.01&0.00&0.20&0.00&0.00&0.00&0.01&0.01\\
DodgerLoopWeekend&0.92&0.53&0.07&-0.04&0.83&-0.01&0.09&0.92\\
Earthquakes&0.00&0.00&0.03&0.00&0.00&0.00&-0.09&-0.01\\
ECG5000&0.51&0.43&0.49&0.71&0.35&0.52&0.62&0.59\\
ECGFiveDays&0.00&0.00&0.40&0.03&0.00&0.22&0.03&0.02\\
ElectricDevices&0.16&0.05&0.09&0.19&0.08&0.00&0.14&0.20\\
EOGHorizontalSignal&0.21&0.20&0.14&0.18&0.18&0.10&0.00&0.22\\
EOGVerticalSignal&0.10&0.11&0.11&0.10&0.09&0.09&0.13&0.08\\
EthanolLevel&0.00&0.00&0.00&0.00&0.00&0.00&0.00&0.00\\
FaceAll&0.22&0.21&0.45&0.26&0.04&0.30&0.14&0.28\\
FaceFour&0.32&0.29&0.42&0.14&0.29&0.48&0.14&0.32\\
FacesUCR&0.21&0.20&0.41&0.24&0.04&0.30&0.14&0.28\\
FiftyWords&0.26&0.24&0.20&0.40&0.09&0.24&0.28&0.31\\
Fish&0.21&0.18&0.27&0.28&0.07&0.28&0.00&0.24\\
FreezerRegularTrain&0.29&0.25&0.28&0.28&0.29&0.27&0.05&0.24\\
FreezerSmallTrain&0.29&0.24&0.28&0.28&0.29&0.27&0.05&0.27\\
Fungi&0.64&0.63&0.15&0.55&0.61&0.85&0.52&0.72\\
GunPoint&-0.01&-0.01&-0.01&-0.01&-0.01&-0.01&-0.01&-0.01\\
GunPointAgeSpan&0.00&0.00&0.00&0.00&0.00&0.00&0.00&0.00\\
GunPointMaleVersusFemale&0.23&0.23&0.00&0.23&0.23&0.23&0.23&0.23\\
GunPointOldVersusYoung&0.24&0.24&0.00&0.24&0.24&0.24&0.24&0.24\\
Ham&0.05&0.03&0.05&0.03&0.04&0.00&0.00&0.06\\
HandOutlines&0.29&0.28&0.32&0.04&0.29&0.01&0.00&0.39\\
Haptics&0.06&0.06&0.06&0.06&0.06&0.08&0.04&0.06\\
Herring&0.00&0.00&0.00&0.00&0.00&-0.01&0.03&0.02\\
HouseTwenty&0.11&0.11&0.11&0.18&0.12&0.16&-0.01&0.07\\
InlineSkate&0.01&0.01&0.04&0.04&0.01&0.01&0.02&0.01\\
InsectEPGRegularTrain&1.00&1.00&0.00&1.00&0.96&1.00&1.00&1.00\\
InsectEPGSmallTrain&0.91&0.91&0.00&1.00&1.00&1.00&1.00&1.00\\
InsectWingbeatSound&0.34&0.33&0.17&0.25&0.14&0.33&0.18&0.33\\
ItalyPowerDemand&0.00&0.35&0.01&0.00&0.00&0.54&0.17&0.00\\
LargeKitchenAppliances&0.02&0.02&0.01&0.03&0.02&0.01&0.06&0.02\\
Lightning7&0.26&0.22&0.35&0.20&0.15&0.23&0.18&0.30\\
Mallat&0.77&0.72&0.70&0.80&0.95&0.58&0.43&0.84\\
Meat&0.62&0.62&0.46&0.55&0.49&0.82&0.45&0.44\\
MedicalImages&0.05&0.04&0.08&0.05&0.05&0.04&-0.04&0.04\\
MelbournePedestrian&0.44&0.45&0.10&0.41&0.43&0.41&0.24&0.47\\
MiddlePhalanxOutlineAgeGroup&0.35&0.34&0.39&0.42&0.42&0.01&-0.03&0.43\\
MiddlePhalanxOutlineCorrect&0.00&0.00&0.00&-0.01&0.00&-0.02&-0.02&-0.01\\
MiddlePhalanxTW&0.37&0.37&0.46&0.58&0.44&-0.01&0.11&0.37\\
MixedShapesRegularTrain&0.44&0.30&0.44&0.47&0.38&0.38&0.13&0.55\\
MixedShapesSmallTrain&0.46&0.40&0.48&0.53&0.41&0.40&0.52&0.55\\
MoteStrain&0.39&0.36&0.61&0.42&0.45&0.55&0.00&0.38\\
NonInvasiveFetalECGThorax1&0.43&0.38&0.33&0.35&0.15&0.12&0.08&0.47\\
NonInvasiveFetalECGThorax2&0.50&0.45&0.46&0.49&0.19&0.22&0.17&0.54\\
OliveOil&0.51&0.40&0.49&0.36&0.70&0.23&0.15&0.63\\
OSULeaf&0.14&0.12&0.24&0.13&0.05&0.07&0.01&0.18\\
PhalangesOutlinesCorrect&0.01&0.01&0.01&0.00&0.01&0.01&0.00&0.00\\
Phoneme&0.02&0.01&0.04&0.01&0.00&0.00&0.01&0.00\\
PigAirwayPressure&0.05&0.04&0.01&0.06&0.06&0.04&0.05&0.05\\
PigArtPressure&0.16&0.14&0.00&0.14&0.09&0.15&0.11&0.19\\
PigCVP&0.07&0.07&0.00&0.09&0.08&0.05&0.04&0.08\\
Plane&0.70&0.63&0.74&0.80&0.86&0.83&1.00&0.80\\
PowerCons&0.73&0.61&0.05&0.66&0.75&0.15&0.00&0.86\\
ProximalPhalanxOutlineAgeGroup&0.42&0.43&0.50&0.57&0.51&0.35&0.02&0.52\\
ProximalPhalanxOutlineCorrect&0.07&0.06&0.07&0.05&0.07&0.06&0.11&0.05\\
ProximalPhalanxTW&0.40&0.40&0.44&0.32&0.38&0.25&0.33&0.42\\
RefrigerationDevices&0.00&0.00&0.01&0.00&0.00&0.00&0.01&0.00\\
Rock&0.22&0.19&0.06&0.23&0.23&-0.01&0.23&0.30\\
ScreenType&0.02&0.01&0.01&0.01&0.03&0.00&0.00&0.02\\
SemgHandGenderCh2&0.00&0.00&0.13&0.00&-0.01&0.01&-0.01&0.00\\
SemgHandMovementCh2&0.14&0.14&0.05&0.16&0.14&0.01&0.00&0.13\\
SemgHandSubjectCh2&0.08&0.08&0.12&0.10&0.07&0.03&0.00&0.10\\
ShapeletSim&0.00&0.01&0.46&0.00&0.00&0.00&0.00&0.00\\
ShapesAll&0.36&0.31&0.36&0.35&0.06&0.12&0.12&0.37\\
SmallKitchenAppliances&0.00&0.03&0.00&0.07&0.00&0.00&0.00&0.00\\
SmoothSubspace&0.44&0.29&0.18&0.43&0.43&0.35&0.03&0.50\\
SonyAIBORobotSurface1&0.34&0.23&0.46&0.71&0.53&0.03&0.00&0.41\\
SonyAIBORobotSurface2&0.32&0.21&0.18&0.30&0.32&-0.03&-0.02&0.26\\
StarLightCurves&0.52&0.35&0.53&0.53&0.52&0.54&0.68&0.51\\
Strawberry&-0.02&0.01&-0.02&-0.01&0.00&-0.04&0.08&-0.05\\
SwedishLeaf&0.30&0.28&0.32&0.15&0.27&0.09&0.04&0.30\\
Symbols&0.64&0.58&0.71&0.65&0.67&0.38&0.82&0.68\\
SyntheticControl&0.59&0.34&0.60&0.64&0.52&0.26&0.31&0.61\\
ToeSegmentation1&0.00&0.00&0.00&0.00&0.00&0.01&0.01&0.02\\
ToeSegmentation2&0.00&0.00&0.27&0.00&0.00&0.02&-0.01&0.05\\
Trace&0.34&0.35&0.32&0.41&0.34&0.34&0.66&0.33\\
TwoLeadECG&0.00&0.00&0.08&0.02&0.00&0.00&0.03&0.00\\
TwoPatterns&0.02&0.02&0.21&0.07&0.02&0.08&0.29&0.02\\
UMD&0.15&0.13&0.14&0.15&0.15&0.12&0.21&0.14\\
UWaveGestureLibraryAll&0.55&0.50&0.62&0.52&0.17&0.54&0.25&0.59\\
UWaveGestureLibraryX&0.34&0.32&0.30&0.39&0.32&0.40&0.49&0.41\\
UWaveGestureLibraryY&0.33&0.30&0.24&0.35&0.30&0.23&0.26&0.34\\
UWaveGestureLibraryZ&0.31&0.29&0.34&0.34&0.31&0.31&0.28&0.29\\
Wine&0.00&0.00&-0.01&-0.01&-0.01&0.00&-0.01&-0.01\\
WordSynonyms&0.16&0.14&0.19&0.23&0.10&0.14&0.18&0.17\\
Worms&0.02&0.00&0.05&0.02&0.01&0.00&0.00&0.07\\
WormsTwoClass&0.00&0.00&0.00&0.00&0.00&0.00&0.00&-0.01\\
Yoga&0.00&0.00&0.00&0.00&0.00&0.00&0.00&0.00\\
\hline%
\end{longtable}%
\end{center}
\end{small}

%% file: spread_table.tex
\begin{table}[ht]
\caption{Pairwise spread of ARI scores between clustering methods.}\label{tbl:spreadscores}
\centering
\resizebox{\columnwidth}{!}{%
\begin{tabular}{|p{4.8cm}|p{1.4cm}|p{1.2cm}|p{1.0cm}|p{0.9cm}|p{1.0cm}|p{1.2cm}|p{1.2cm}|p{1.2cm}|}
\hline
Clustering method & Agglo-merative (Euc) & K-means (DTW) & K-means (Euc) & C-means (Euc) & K-med (Euc) & K-means (shape) & Density peaks (Euc) & Density Peaks (DTW)\\
\hline
Agglomerative (Euclidean)& - &0.020&0.004&0.011&0.011&0.050&0.043&0.054\\
K-means (DTW)& - & - &0.016&0.025&0.017&0.041&0.043&0.052\\
K-means (Euclidean)& - & - & - &0.010&0.005&0.043&0.036&0.045\\
C-means (Euclidean)& - & - & - & - &0.011&0.053&0.038&0.043\\
K-medoids (Euclidean)& - & - & - & - & - &0.042&0.021&0.032\\
K-means (shape-based)& - & - & - & - & - & - &0.060&0.067\\
Density Peaks (Euclidean)& - & - & - & - & - & - & - &0.021\\
Density Peaks (DTW) & - & - & - & - & - & - & - & -\\
\hline
\end{tabular}%
}
\end{table}